\documentclass[10pt,conference]{IEEEtran}

\usepackage{graphicx} % Required for inserting images

\usepackage{amsmath}
\usepackage{tikz}
\usepackage{url}

\usetikzlibrary{calc,backgrounds,arrows.meta, fit}

\pgfdeclarelayer{bg}
\pgfsetlayers{bg,main}

\definecolor{semcol}{RGB}{124,32,114}     % semantic node purple
\definecolor{epicol}{RGB}{0,150,170}      % episodic node teal
\definecolor{epifill}{RGB}{198,238,245}   % episodic fill
\definecolor{bgpurple}{RGB}{244,236,250}  % big background disk
\definecolor{qorange}{RGB}{225,135,35}    % query box border
\definecolor{qfill}{RGB}{255,238,212}     % query box fill
\definecolor{tracecol}{RGB}{20,30,200}    % retrieved trace blue
\definecolor{glow}{RGB}{255,205,110}      % glow behind SM6

\title{Escaping Python Dependency Hell: A Hybrid Replay-and-Repair Pipeline for Python Dependency Resolution}
\IEEEoverridecommandlockouts

\date{June 2026}

\author{
    \IEEEauthorblockN{
        Veronica Poweska\IEEEauthorrefmark{1}\textsuperscript{$\diamond$},
        Ariana Oyanguren\IEEEauthorrefmark{1}\textsuperscript{$\diamond$},
        Jessica Pourleyli\IEEEauthorrefmark{1}\textsuperscript{$\diamond$},
        Sourena Khanzadeh
        \IEEEauthorrefmark{2}\IEEEauthorrefmark{3},
                \thanks{Sourena Khanzadeh's affiliation with Flybits began after the work reported in this paper had commenced.}
        Manar Alalfi\IEEEauthorrefmark{1}}
    \IEEEauthorblockA{\IEEEauthorrefmark{1}Toronto Metropolitan University, Computer Science, Toronto, Ontario, Canada\\
        \{vpoweska, aoyangurenvaldivia, jessica.pourleyli, sourena.khanzadeh, manar.alalfi\}@torontomu.ca}
    \IEEEauthorblockA{\IEEEauthorrefmark{2}Toronto Metropolitan University, Creative School, Toronto, Ontario, Canada}
    \IEEEauthorblockA{\IEEEauthorrefmark{3}Flybits, Creative AI Hub, Toronto, Ontario, Canada\\
        sourena.khanzadeh@flybits.com}
    \IEEEauthorblockA{\textsuperscript{$\diamond$}These authors contributed equally to this work.}
}

\begin{document}

\maketitle

\begin{abstract} Dependency conflicts in Python ecosystems arise from incompatible version constraints, missing packages, and undocumented compatibility relationships, causing many real-world code snippets to fail at execution. This paper presents PLLM+, a hybrid dependency-repair pipeline evaluated on the HG2.9K benchmark of 2,891 dependency-failing snippets. PLLM+ prioritizes inexpensive deterministic steps before invoking LLM-based repair: static AST-based interpreter inference, replay of historically successful dependency configurations from the competition-provided solutions database, and live PyPI validation of candidate package versions. When these steps do not resolve a case, the system falls back to a structured LLM-based repair loop with typed error classification and Proposer/Critic agents. On HG2.9K, PLLM+ solves 1,500 out of 2,891 snippets, compared with 1,169 solved by the PLLM baseline. It also reduces average runtime from 368.7 to 71.8 seconds per snippet. Most successful fixes come from replaying known configurations: 1,495 of the 1,500 successful fixes are produced by the solutions database, while the LLM fallback accounts for 5 additional fixes. These results suggest that, in this benchmark setting, deterministic reuse of previously validated dependency configurations is a simple and effective strategy, with LLM-based repair serving as a secondary fallback for cases not covered by prior solutions. The code can be found at \footnote{\url{https://github.com/vpoweska/fse-aiware-python-dependencies}} and will be released upon publication. \end{abstract}

\section{Introduction}

Modern software engineering depends heavily on third-party libraries, but this
reliance introduces a persistent problem: dependency conflicts. Incompatible
version constraints cause installation failures and broken execution
environments, and prior work has shown that a substantial portion of real-world
Python programs fail to run due to missing or incompatible dependencies
\cite{horton2018gistable}. The problem has grown as packages drop support for
older interpreters and change their constraints over time \cite{kula2018updates}.
The FSE-AIWare codebase formalizes this challenge, 
building agentic systems that automatically repair Python dependency
specifications on the HG2.9K dataset; it provides a reference implementation
(PLLM) and its evaluation artifacts, and encourages reuse of previously
successful configurations \cite{fse2026competition, horton2018gistable}.

Resolving conflicts automatically is hard because compatibility relationships are largely implicit, the space of valid combinations is large, and fixing one conflict can introduce another. Existing LLM-based approaches tackle this with iterative error-feedback loops \cite{bartlett2025last}, but can still generate dependencies or versions that look plausible but fail in practice. We introduce \textbf{PLLM+}, a hybrid pipeline that first reuses historical solution data, performs static syntax-based interpreter inference, and validates candidate versions against PyPI before falling back to LLM-based repair. Our goal is not to replace deterministic dependency resolution with LLM reasoning, but to study how far a replay-first strategy can go on the HG2.9K benchmark and where an LLM fallback can add coverage. We address three research questions: whether PLLM+ improves the success rate over PLLM on HG2.9K (RQ1), whether it resolves conflicts more efficiently in average time per snippet (RQ2), and what share of its fixes come from the solutions database versus the LLM fallback (RQ3). Our contributions are: \begin{itemize} \item We present PLLM+, a hybrid dependency-repair pipeline that prioritizes deterministic replay and validation before invoking LLM-based repair. \item We evaluate PLLM+ on the HG2.9K benchmark and show that it improves success rate and runtime compared with the PLLM baseline under the benchmark setting. \item We report where successful fixes come from, showing that the solutions database is the primary source of successful repairs, while the LLM fallback provides limited additional coverage. \item We provide a reproducible implementation integrated with the FSE-AIWare framework. The code and artifacts will be released upon publication.
\end{itemize}

\begin{figure*}
    \centering
    \includegraphics[width=1\linewidth]{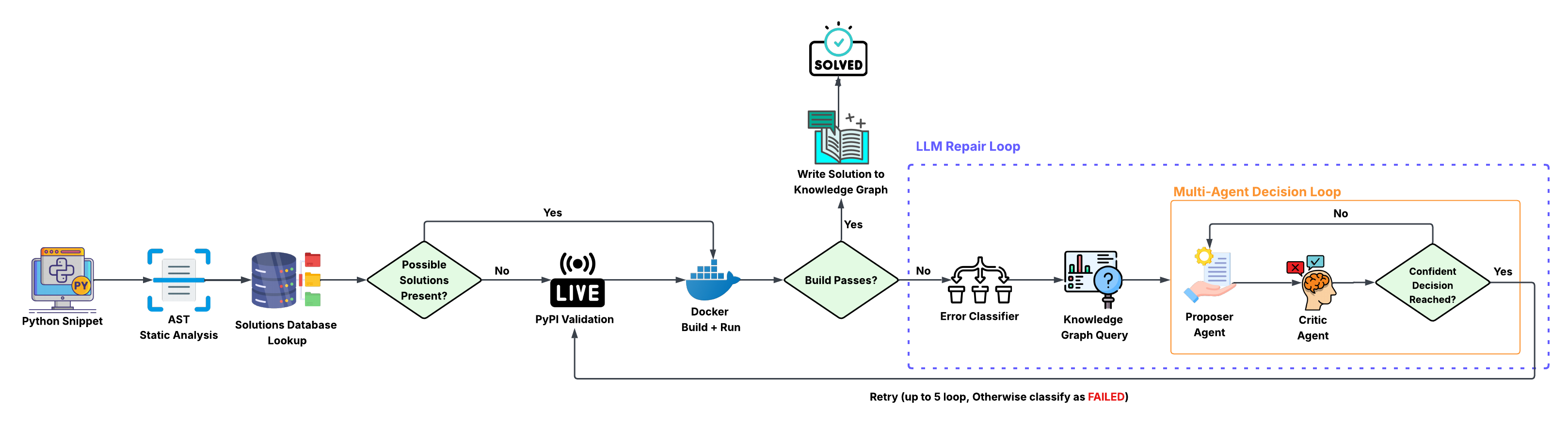}
    \caption{High-Level Architecture of the PLLM+ enhanced dependency repair pipeline.}
    \label{fig:architecture}
\end{figure*}

\section{Related Work}

Early dependency-inference systems treat the task as knowledge retrieval.
DockerizeMe builds an inter-dependency graph to generate a Dockerfile for a
snippet \cite{horton2019dockerizeme}, and PyEGo extends this with a knowledge
graph that jointly infers packages, interpreter, and system libraries at
compatible versions \cite{ye2022knowledge}; later graph-based tools such as
ReadPy add iterative refinement and adaptive knowledge maintenance
\cite{10373775}. These methods are reproducible but bounded by the coverage and
freshness of a hand-built knowledge base \cite{kula2018updates}, a difficulty
compounded by the scale and prevalence of configuration issues across the PyPI
ecosystem \cite{peng2024less}. PLLM instead casts repair as LLM generation,
using retrieval-augmented prompting with an iterative build-feedback loop,
though it remains susceptible to hallucinated versions \cite{bartlett2025last}.
PLLM+ keeps deterministic retrieval as a fast path while validating every
candidate against live PyPI to eliminate non-existent releases before any build.

More broadly, LLM-based program repair spans prompting, procedural, and agentic
designs, the last delegating loop control to the model at higher latency and
cost \cite{yang2025survey, bouzenia2025repairagent}. Cooperative multi-agent
frameworks extend this further, coordinating specialised agents across software
development tasks \cite{khanzadeh2025agentmesh} and refining inter-agent
communication \cite{du2026learning}. PLLM+ instead adopts a procedural design
with the LLM as a bounded fallback, and for that fallback uses multi-agent
debate, which improves factual accuracy over single-model reasoning that tends
to confirm its own errors \cite{du2024multiagent}. Our Proposer--Critic loop
instantiates this for dependency repair, with all surviving candidates still
validated against PyPI.

\section{Methodology}

Figure~\ref{fig:architecture} presents an overview of the PLLM+ architecture.
Given a single Python snippet, the goal is to construct a Docker container in
which it executes without dependency-related failures. The pipeline proceeds
through four stages in priority order, stopping as soon as a solution is found,
and is deliberately ordered so that cheap, deterministic operations are
attempted before LLM reasoning is invoked as a fallback.

\paragraph{\textbf{Stage 1: Static AST Analysis}}
The snippet is parsed with Python's built-in \texttt{ast} module to determine
the minimum required interpreter version \cite{python_ast_docs}, inspecting the
syntax tree for version-specific constructs (e.g., the walrus operator
\texttt{:=}, valid only in Python~3.8 or later \cite{python38_whatsnew}). This
establishes a hard lower bound before any LLM call, preventing the system from
proposing or testing versions that are syntactically incompatible with the
snippet and steering the pipeline toward versions more likely to succeed.

\paragraph{\textbf{Stage 2: Solutions Database}}
The competition provides, and encourages reuse of, result archives from the
original PLLM experiment together with PyEGo and ReadPy artifacts
\cite{fse2026competition}. We parse these tar archives of YAML results for all
2,891 snippets and, for each successful run, record the exact package versions
from the final iteration before the success marker. Multiple passing
configurations per snippet are preserved under unique run identifiers rather
than overwritten. At query time, all known configurations for a snippet are
returned in ascending order of package count, and each is built and run until
one succeeds or all are exhausted. This exhaustive replay means that even when
the first solution uses an incompatible interpreter or a package version that
has since changed on PyPI, later solutions are tried automatically without
invoking the LLM. A compatibility filter using the Stage~1 lower bound discards
configurations that will always fail, avoiding wasted Docker builds.

\paragraph{\textbf{Stage 3: Live PyPI Validation}}
When no database solution succeeds, the system falls through to the LLM
pipeline. Beforehand, candidate versions are validated against the PyPI JSON API
\cite{pypi_json_api}: for each detected import, the resolver queries PyPI for the
real list of versions available for the target Python version and returns the
latest compatible release. A hard-coded cap table handles Python~2.7, since many
packages dropped 2.7 support at different releases after its end-of-life
\cite{numba_python27_eol}, creating compatibility boundaries not always captured
by PyPI metadata \cite{pyradar}. This eliminates hallucinated or non-existent
versions before they reach a build, a major source of wasted attempts in the
PLLM baseline.

\paragraph{\textbf{Stage 4: Multi-Agent Repair Loop}}
On a failed build, the error is classified into a structured type
(\texttt{ModuleNotFoundError}, \texttt{ImportError}, \texttt{DependencyConflict},
\texttt{NonZeroCode}, or \texttt{SyntaxError}), each mapped to a specialist
prompt that injects the most relevant context; for \texttt{NoMatchingDistribution},
for instance, the available versions parsed from the pip error are supplied so
the model selects from versions known to exist rather than guessing
\cite{feuer2025judgment}. Repair then uses a two-role debate
\cite{du2024multiagent}: a single instance that both generates and evaluates a
candidate tends to confirm its own reasoning and repeat the same incorrect
configuration, whereas separating generation from critique exploits the finding
that models detect flaws more reliably than they avoid them. The Proposer
produces a revised package set from the error class, current configuration, and
knowledge graph context; the Critic independently screens it for implausible
versions, missing dependencies, and known incompatibilities before a build is
triggered. Every surviving candidate is re-validated against PyPI, so
hallucinated versions cannot enter the pipeline regardless of what the agents
propose. The loop repeats for up to five iterations.

\paragraph{\textbf{Knowledge Graph}}
Across a batch run, verified package combinations accumulate in a persistent
knowledge graph indexed by package name and Python version. When the LLM
pipeline runs, matching entries for the snippet's imports are injected as
retrieved context. This both narrows the search space by anchoring suggestions
to combinations already proven to work and produces a compounding effect:
solutions from early snippets benefit later ones sharing common dependencies,
which is especially effective for frequent packages such as \texttt{numpy},
\texttt{pandas}, and \texttt{requests}.

\subsection{Validation and Optimization}
A configuration is valid only if the container builds and the snippet executes
without dependency-related failures, the same build-and-run criterion used by
the competition and PLLM, enabling a direct CSV-to-CSV comparison on snippet ID
and outcome \cite{du2024multiagent}. Three optimizations reduce wasted effort.
Exhaustive replay (above) is the most impactful, since a snippet may have been
solved on a different Python version across PLLM runs, so testing all stored
configurations before the LLM conserves resources. Second, a
\texttt{from versions: none} error---indicating a package absent from PyPI---is
detected directly from pip output, and the offending package is dropped without
entering the debate loop. Third, duplicate database entries are detected via a
content hash of the package set and skipped, avoiding repeated failing builds.

\subsection{Implementation}
PLLM+ is implemented in Python as separable components---static analysis,
solutions database management, PyPI validation, error classification,
multi-agent debate, and persistent knowledge storage---interacting through
well-defined interfaces. Separating deterministic from LLM-assisted stages makes
behaviour on database-covered snippets fully reproducible and independent of
model non-determinism while retaining LLM flexibility for novel failures. The
system is compatible with the FSE-AIWare framework and accepts the same input
format as PLLM, allowing direct comparison on HG2.9K.

\section{Evaluation}
In this section, we describe the setup used to run the experiments and provide a detailed discussion of empirical results obtained for research questions RQ1-RQ3.

\subsection{Experimental setup}
All experiments were run on a GPU cluster, each machine uses an 11th Gen Intel Core i7 processor with 8 cores and 16 threads, 15 GB of RAM, and two NVIDIA GeForce GTX 1070 GPUs. The machines run Ubuntu 24.04.

Our system uses an Ollama server to run the language model, specifically Gemma 2. All parts of the system were deployed using Docker containers so that the environment stayed consistent across runs. The containers communicated through a shared Docker network. To evaluate the tool, we ran it on the full set of final Python code snippets in the dataset. Each snippet was processed once. 

The results from every run were saved automatically in a file called \textit{summary\_all\_runs.csv}. This file includes the fields: \textit{snippet\_id, snippet\_path, return\_code, solved, elapsed\_seconds, status, and error}.

Here, \textit{return\_code} shows whether the run finished properly, solved shows if the snippet was handled successfully, and \textit{elapsed\_seconds} shows how long it took. The status field gives a short result label, and error stores the failure message if something went wrong.

To make the experiments reproducible, the code, setup steps, and instructions are all included in the project repository.

\subsection{Results}
\subsubsection{RQ1: Does PLLM+ achieve a higher success rate than the
PLLM baseline on the HG2.9K benchmark?} \mbox{}

To answer this research question, we compare how many snippets were
successfully solved by PLLM and PLLM+ on HG2.9K using the
\textit{summary\_all\_runs.csv} files from both tools.

\begin{figure}[ht]
    \centering
    \includegraphics[width=1\linewidth]{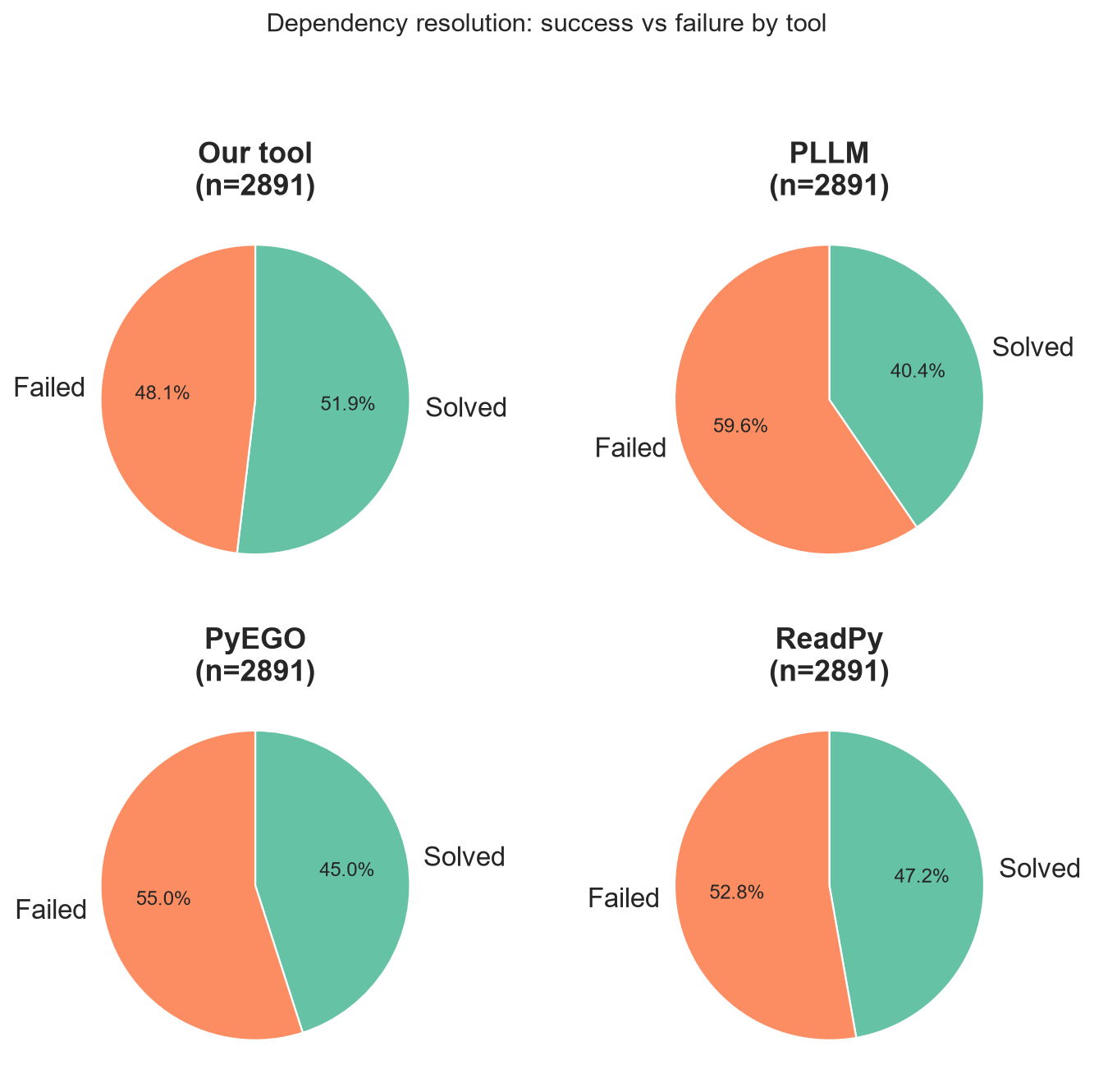}
    \caption{Success rate across tools}
    \label{fig:success_rate}
\end{figure}

As shown in Figure~\ref{fig:success_rate}, PLLM solved 1,169 out of 2,891
snippets, while PLLM+ solved 1,500 out of 2,891. This gives PLLM a success rate
of 40.4\% and PLLM+ a success rate of 51.9\%. PLLM+ solved 331 more snippets
than the baseline and improved the success rate by 11.5 percentage points. Based
on these results, PLLM+ performs better than PLLM on the benchmark, suggesting
that the replay-based approach helps the system resolve more dependency issues
successfully.

\subsubsection{RQ2: Does PLLM+ resolve dependency conflicts more efficiently than the PLLM baseline, measured by average resolution time per snippet?} \mbox{}

\begin{table}[htbp]
\centering
\small
\caption{Failure categories: counts per tool on unsuccessful runs (same 2891 snippets). Our tool uses \texttt{status}; baselines use \texttt{result}.}
\label{tab:failure-categories-comparison}
\resizebox{\linewidth}{!}{
\begin{tabular}{lrrrrr} 
\hline
Category & Our tool & PLLM & PyEGO & ReadPy & Total \\
\hline
ImportError & 0 & 433 & 707 & 315 & 1455 \\
failed & 1374 & 0 & 0 & 0 & 1374 \\
ModuleNotFound & 0 & 8 & 634 & 492 & 1134 \\
Build failure & 0 & 0 & 0 & 612 & 612 \\
SyntaxError & 0 & 494 & 0 & 58 & 552 \\
OtherFailure & 0 & 45 & 247 & 0 & 292 \\
NoMatchingDistribution & 0 & 282 & 0 & 1 & 283 \\
OtherPass & 0 & 222 & 0 & 0 & 222 \\
AttributeError & 0 & 83 & 1 & 48 & 132 \\
CouldNotBuildWheels & 0 & 83 & 0 & 0 & 83 \\
TypeError & 0 & 28 & 0 & 0 & 28 \\
InvalidRequirement & 0 & 17 & 0 & 0 & 17 \\
timeout & 17 & 0 & 0 & 0 & 17 \\
NameError & 0 & 12 & 0 & 0 & 12 \\
FailedToRun & 0 & 9 & 0 & 0 & 9 \\
DjangoPass & 0 & 6 & 0 & 0 & 6 \\
\hline
\end{tabular}
}
\end{table}

To answer this research question, we compare both the average and median runtime per snippet for PLLM and PLLM+ using the \textit{elapsed\_seconds} values recorded in the output CSV files.

PLLM had an average runtime of 368.74 seconds per snippet, while PLLM+ averaged
71.77 seconds---296.97 seconds faster, or about 5.1 times faster than the
baseline. The same gap holds at the median: 364.48 seconds for PLLM versus 21.5
seconds for PLLM+. 

One likely explanation is that many PLLM+ snippets were resolved quickly through replaying known dependency configurations, while more difficult cases required additional fallback processing. Overall, these results show that PLLM+ resolves dependency issues much more efficiently than PLLM.
\subsubsection{RQ3: What portion of the successful fixes arise from configurations stored in the competition-provided solutions database
compared to the LLM-based fallback pipeline?} \mbox{}

To answer this research question, we examined how each successful fix in PLLM+ was produced. Specifically, we tracked whether a solved snippet was resolved through replaying a stored dependency configuration from the solutions database or through the LLM-based fallback pipeline.

\begin{figure}[ht]
    \centering
    \includegraphics[width=1\linewidth]{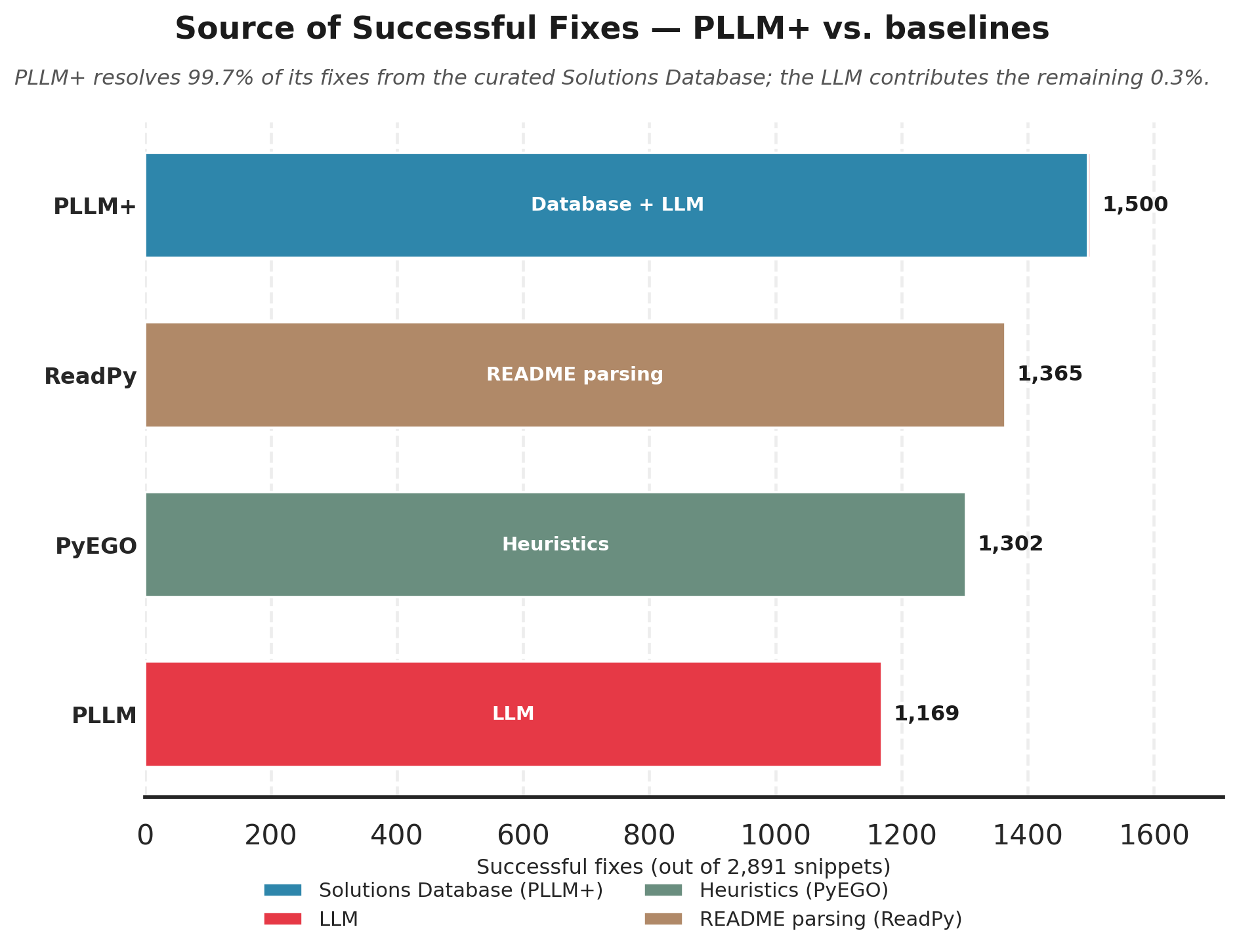}
    \caption{Source of successful fixes in PLLM+: solutions database replay vs.\ LLM fallback.}
    \label{fig:source_of_fix}
\end{figure}

As shown in Figure~\ref{fig:source_of_fix}, PLLM+ successfully solved 1500 snippets in total. Out of these, 1495 were solved using the solutions database, while only 5 were solved using the LLM fallback pipeline. This corresponds to 99.7\% of successful fixes coming from database replay and 0.3\% coming from the LLM.
These results show that almost all successful fixes in PLLM+ came from reusing previously known working configurations rather than generating new solutions through the LLM. This suggests that the main strength of PLLM+ comes from its ability to retrieve and replay valid dependency setups efficiently.

This is an important result because it shows that many dependency resolution problems in the benchmark can be handled by matching and reusing known configurations instead of resolving them from scratch. In practice, this makes the system both faster and more reliable.

At the same time, the LLM fallback still plays a useful role for cases where no matching configuration is found in the database. Even though it contributed to only a small number of successful fixes in our experiments, it provides coverage for cases that cannot be solved through direct reuse alone.

Overall, these results suggest that the knowledge-base component is the primary driver of PLLM+ performance, while the LLM fallback acts as a secondary recovery mechanism.

Table~\ref{tab:failure-categories-comparison} breaks down the unsuccessful runs
by failure category across all four tools on the same 2,891 snippets. The
knowledge-graph baselines fail in semantically specific ways---PyEGo and ReadPy
together account for the bulk of \texttt{ImportError} and \texttt{ModuleNotFound}
cases, and ReadPy additionally incurs 612 build failures---while PLLM's failures
spread across \texttt{SyntaxError}, \texttt{NoMatchingDistribution}, and other
runtime categories. In contrast, PLLM+ records its unsolved cases almost entirely
as a single \texttt{failed} status (1,374) plus 17 timeouts, reflecting that our
pipeline reports a coarse build-and-run outcome rather than classifying the
underlying error on unsolved snippets.

\section{Discussion}
\begin{figure*}
    \centering
    \includegraphics[width=1\linewidth]{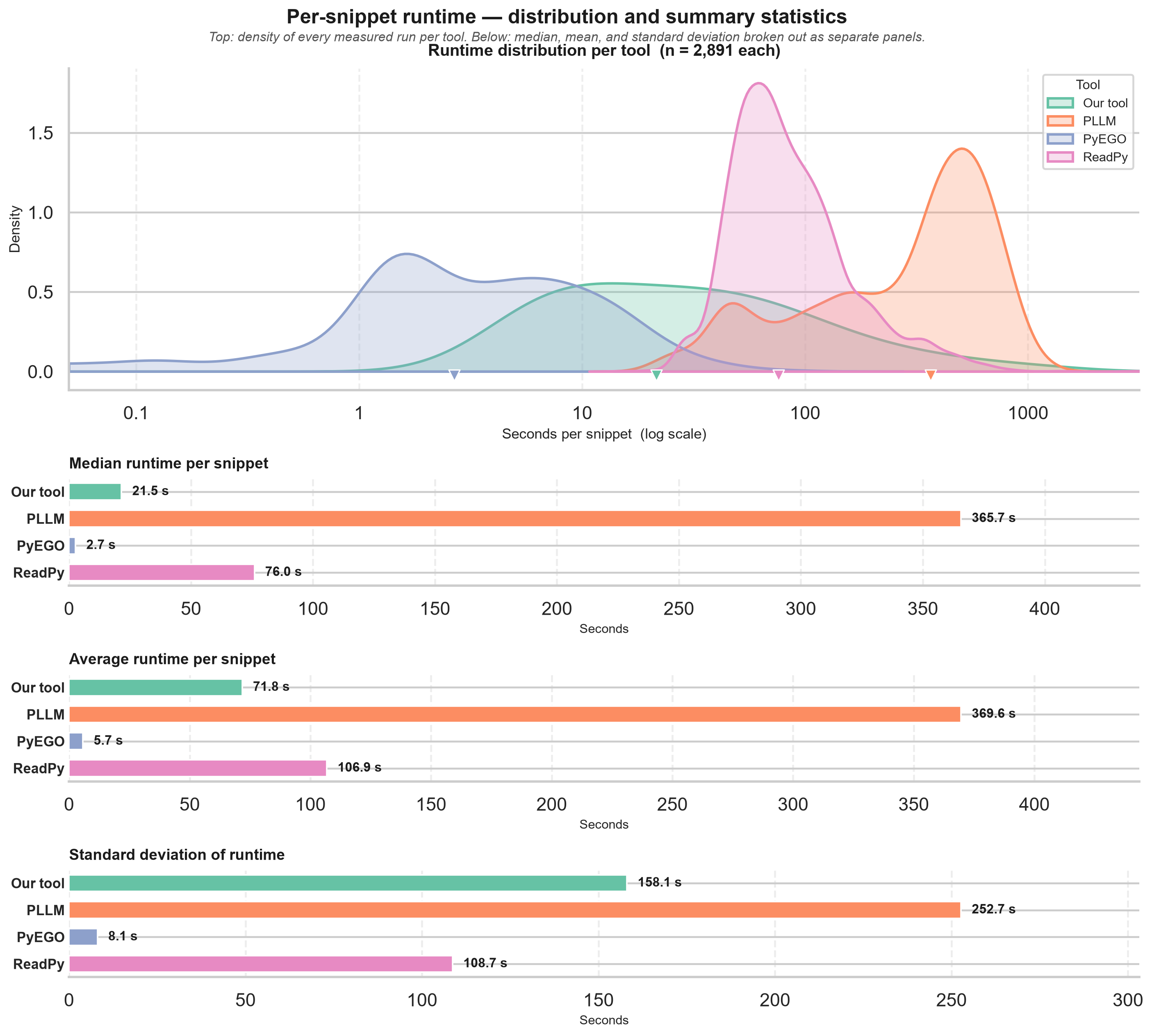}
    \caption{Runtime Distribution Across Different Tool Sets}
    \label{fig:runtime}
\end{figure*}

Figure~\ref{fig:runtime} shows the full per-snippet runtime
distribution for both tools, making this separation visible across the benchmark
rather than only in aggregate.

\paragraph{\textbf{Implications.}}
The main lesson from PLLM+ is that dependency repair can benefit substantially
from simple deterministic reuse before invoking more expensive LLM-based
reasoning. On HG2.9K, many failing snippets can be repaired by replaying
dependency configurations that were previously observed to work. This makes the
repair process faster, more reproducible, and less dependent on model behavior.

The LLM component remains useful as a fallback mechanism, but our results show
that it is not the main source of successful fixes in this evaluation. Instead,
the strongest performance comes from ordering the pipeline by cost: first
applying static interpreter inference, then replaying known configurations, then
validating candidates against PyPI, and only then invoking LLM-based repair.
This suggests that LLMs may be most useful in dependency repair when used
selectively, after cheaper sources of evidence have been exhausted.

\subsection{Threats to Validity}
\paragraph{\textbf{Internal.}} Repair depends on classifying build failures into
error types via string matching over common patterns (version conflict, no
matching distribution, missing system dependency, Python version mismatch, and
module not found); a misclassification can trigger the wrong strategy, and design
choices such as version filtering or package removal may bias the process by
excluding valid configurations. We mitigate this with structured error categories
and consistent filtering and validation rules across all runs.

\paragraph{\textbf{External.}} Results are bounded by HG2.9K, a specific set of
GitHub snippets that may not represent all real-world dependency issues. Crucially,
most successful fixes draw on the provided solutions database, which is built from
the same dataset, so performance on completely new or unseen data may differ. We
reduce this threat by evaluating on the full benchmark under the same setup as the
PLLM baseline.

\subsection{Limitations and Future Work}
PLLM+'s main limitation is its reliance on the solutions database: high success
and low runtime depend on how many cases are already covered. The database is
currently limited to benchmark-provided configurations and could be expanded with
external sources---open-source repositories, package documentation, or developer
discussions such as Stack Overflow---to broaden coverage.

\section{Conclusion}

This paper presented PLLM+, a hybrid pipeline for automated Python dependency repair. The system combines static interpreter inference, replay of previously validated dependency configurations, live PyPI validation, and a bounded LLM-based fallback. On the HG2.9K benchmark, PLLM+ solved 1,500 out of 2,891 snippets, compared with 1,169 solved by the PLLM baseline, while substantially reducing average runtime. The evaluation shows that the main source of improvement is deterministic replay of known-good configurations. In our experiments, 1,495 of the 1,500 successful fixes came from the solutions database, while the LLM fallback solved 5 additional cases. This result suggests that, at least for this benchmark, dependency repair benefits strongly from efficient reuse and validation of prior solutions. We therefore view PLLM+ as a practical hybrid design: use deterministic methods whenever possible, and reserve LLM-based repair for cases where prior validated configurations are unavailable. A natural next step is to evaluate the same design on held-out snippets or external repositories with less overlap with the existing solutions database. Future work should also strengthen the LLM fallback and expand the retrieval source beyond benchmark artifacts, for example by mining open-source repositories, package documentation, and developer discussions for validated dependency configurations.
\bibliographystyle{ieeetr}
\bibliography{references}

\end{document}